\documentclass[times,twocolumn,final]{elsarticle}

\usepackage{medima}
\usepackage{framed,multirow}

\usepackage{amssymb}
\usepackage{latexsym}

\usepackage{url}
\usepackage{xcolor}
\usepackage{bm}
\usepackage{hyperref}
\newcommand{\x}{\bm{x}}
\newcommand{\y}{\bm{y}}
\newcommand{\z}{\bm{z}}
\newcommand{\n}{\bm{n}}
\newcommand{\e}{\bm{e}}
\newcommand{\uu}{\bm{u}}
\newcommand{\vv}{\bm{v}}

\newcommand{\xib}{\bm{\xi}}
\usepackage{bm}
\usepackage{amsmath}

\newtheorem{assump}{Assumption}
\newtheorem{rmk}{Remark}[section]
\newtheorem{thm}{Theorem}[section]
\newtheorem{defn}{Definition}[section]
\newtheorem{lem}{Lemma}[section]
\definecolor{newcolor}{rgb}{.8,.349,.1}

\journal{Medical Image Analysis}

\begin{document}

\verso{Cui \textit{et~al.}}

\begin{frontmatter}

\title{K-UNN: $k$-Space Interpolation With Untrained Neural Network}%

\author[1]{Zhuo-Xu \snm{Cui}}
\author[2]{Sen \snm{Jia}}
\author[1]{Qingyong \snm{Zhu}}
\author[2]{Congcong \snm{Liu}}
\author[2]{Zhilang \snm{Qiu}}
\author[3]{Yuanyuan \snm{Liu}}
\author[2]{Jing \snm{Cheng}}
\author[2]{Haifeng \snm{Wang}}
\author[2]{Yanjie \snm{Zhu}}
\author[1,2]{Dong \snm{Liang}\corref{cor1}}
\ead{dong.liang@siat.ac.cn}

\cortext[cor1]{Corresponding author:}

\address[1]{Research Center for Medical AI, Shenzhen Institutes of Advanced Technology, Chinese Academy of Sciences, Shenzhen, China}
\address[2]{Paul C. Lauterbur Research Center for Biomedical Imaging, Shenzhen Institutes of Advanced Technology, Chinese Academy of Sciences, Shenzhen, China.}
\address[3]{National Innovation Center for Advanced Medical Devices, Shenzhen, Guangdong, China.}

\received{1 May 2013}
\finalform{10 May 2013}
\accepted{13 May 2013}
\availableonline{15 May 2013}
\communicated{S. Sarkar}

\begin{abstract}
Recently, untrained neural networks (UNNs) have shown satisfactory performances for MR image reconstruction on random sampling trajectories without using additional full-sampled training data. However, the existing UNN-based approach does not fully use the MR image physical priors, resulting in poor performance in some common scenarios (e.g., partial Fourier, regular sampling, etc.) and the lack of theoretical guarantees for reconstruction accuracy. To bridge this gap, we propose a safeguarded $k$-space interpolation method for MRI using a specially designed UNN with a tripled architecture driven by three physical priors of the MR images (or $k$-space data), including sparsity, coil sensitivity smoothness, and phase smoothness. We also prove that the proposed method guarantees tight bounds for interpolated $k$-space data accuracy. Finally, ablation experiments show that the proposed method can more accurately characterize the physical priors of MR images than existing traditional methods. Additionally, under a series of commonly used sampling trajectories, experiments also show that the proposed method consistently outperforms traditional parallel imaging methods and existing UNNs, and even outperforms the state-of-the-art supervised-trained $k$-space deep learning methods in some cases.
\end{abstract}

\begin{keyword}
\MSC 41A05\sep 41A10\sep 65D05\sep 65D17
\KWD $k$-space interpolation\sep untrained neural networks\sep  theoretical bound\sep  parallel imaging
\end{keyword}

\end{frontmatter}


\section{Introduction}
\label{sec:introduction}
MRI is a widely used imaging modality in routine clinical practice due to its noninvasion, nonionization and excellent visualization of soft-tissue contrast. However, it has also been traditionally limited by its slow data acquisition speed in many applications.
How to reduce the imaging time has become a research hotspot.
In 2006, a high-profile method, termed Compressed Sensing (CS), was proposed (\cite{Candes2006Robust,Donoho2006Compressed}), which theoretically shows that sparse signals can be completely recovered from incomplete measurement under certain conditions. Leveraging the sparse nature of MR images under certain transforms or dictionaries, CS has been successfully applied to accelerated MRI (CS-MRI) (\cite{Lustig2007Sparse}). However, further acceleration is limited due to the complexity limitations of the CS-MRI model. Inspired by the tremendous success of deep learning (DL),
many researchers have committed to applying DL to MR reconstruction and received significant performance gains (\cite{7493320,yang2016deep,zhu2018Image,8756028,8962949,Huang2021Deep,9481093,9481108}). Nevertheless, most of these methods typically require many full-sampled training dataset, which is impractical in the clinic.

Untrained neural networks (UNNs) are intersection models between DL and CS. It not only inherits the powerful representation ability of deep neural networks but also requires no additional training data like CS (\cite{Ulyanov_2018_CVPR}). However, a suitable network architecture is difficult to design and significantly impacts the performance of UNNs. More specifically, instead of relying on sparse nature, UNN captures the prior of the sought solution by parameterizing it through a carefully designed deep neural network. Then, let the solution represented by the network satisfy the data consistency, transforming the sparse constrained optimization problem in CS into an unconstrained network fitting problem. In a nutshell, UNN is regularized by the network architecture (\cite{Dittmer2020Regularization}). With the advantages of UNNs, this field of study has emerged as a competitive method for solving inverse problems (\cite{8581448,9442767,9488215,qayyum2021untrained}). Recently, an image-domain decoder architecture-based UNN (I-UNN) was applied to the MR image reconstruction problem and achieved satisfactory performance on 4x random sampling trajectories (\cite{9488215}).

However, I-UNN cannot perform well in other common sampling scenarios, such as partial Fourier, regular trajectories, etc.
The main reason for this is the \textbf{physical prior underutilization}. In the existing I-UNN methods, the network architecture design mainly relies on computer vision and does not consider the physics prior to MR image. It not only leads to poor reconstruction quality but also the \textbf{lack of theoretical guarantee} for the reconstruction accuracy. It is worth noting that although the literature (\cite{heckel2020compressive}) pointed out that the reconstruction error of I-UNN can be effectively bounded if the coding matrix meets the sub-Gaussian assumption, most of the time, the sub-Gaussian assumption cannot be met in MRI.

\subsection{Contributions and Observations}
Motivated by the abovementioned problems, this paper will tackle the problem that suitable network architectures are difficult to design by guiding the UNN through the physical priors of MR images. Specifically, our work's main contributions and observations are summarized as follows.

\begin{itemize}
	\item For the $k$-space interpolation problem, a tripled UNN architecture is proposed, one for sparse prior to the MR image (or linear predictability of $k$-space data), one for smooth prior to the MR image phase, and one for smooth prior to the coil sensitivities. It is worth mentioning that the proposed tripled UNN is a very flexible framework. If a prior is not satisfied in some cases, the remaining two prior-based modules can form a double UNN architecture.
	\item We prove that the $k$-space data interpolated by the proposed tripled UNN enjoys a tight complexity guarantee to approximate the full-sampled $k$-space data on random and deterministic (including regular and partial Fourier) sampling trajectories.
\item In terms of priors characterization, ablation experiments show that the proposed method can more accurately characterize the physical priors
of MR images than traditional methods.
\item In terms of reconstruction accuracy, experiments on a series of commonly used sampling trajectories show that the proposed tripled UNN consistently outperforms existing UNN-based and traditional parallel imaging methods and even outperforms the state-of-the-art supervised-trained DL-based methods in some cases.
\end{itemize}

The remainder of the paper is organized as follows. Section \ref{sect2} provides some notations and preliminaries. Section \ref{sect_rw} reviews some related work. Section \ref{sect3} discusses the proposed $k$-space interpolation method and its corresponding theoretical guarantees. The implementation details are presented in Section \ref{sect4}. Experiments performed on several data sets are presented in Section \ref{sect5}. The discussions are presented in Section \ref{sect6}. The last section \ref{sect7} gives some concluding remarks. All the proofs are presented in the Appendix.

\section{Related Work}\label{sect_rw}
\subsection{CS $\&$ UNN}
Consider the following general inverse problem $$Ax=y$$
where $A$ is the ill-condition coding operator, $y$ is the measurement data, and $x$ is the variable to be solved. If $x$ satisfies sparsity under a certain sparse transformation $\Psi$, CS can obtain the unique true solution to the inverse problem by solving the following variational problem:
$$\min_{\xi}\frac{1}{2}\|A\Psi(\xi)-y\|^2+\lambda\|\xi\|_1$$
where $\|\cdot\|_1$ is a sparse promoting regularizer. In particular, the unique true solution $x^*=\Psi(\xi^*)$, where $\xi^*$ is the optimal solution of the above variational problem. Instead of relying on sparse nature, UNN captures the prior of the sought solution by parameterizing it through a carefully designed deep neural network $\Phi$, i.e., $x=\Phi(\zeta)$, where $\zeta$ is a random low-dimensional vector. Then, CS can be generalized to
$$\min_{\Phi}\frac{1}{2}\|A\Phi(\zeta)-y\|^2.$$
In a nutshell,  CS fixes the sparse transform $\Psi$ and seeks a sparse coefficient $\xi$ to represent the desired solution, while UNN fixes a sparse (low-dimensional) coefficient $\zeta$ and seeks an adaptive transformation $\Psi$ to represent the desired solution. Namely, the sparse prior in CS is generalized to be implicitly extracted by UNN network architecture.

Existing UNN methods are mainly oriented toward computer vision. Based on the discovery that Encoder-decoder architecture has high impedance to noise and low impedance to the desired image, \cite{Ulyanov_2018_CVPR} applied the UNN with Encoder-decoder architecture to various linear inverse problems, including denoising, inpainting, super-resolution, etc.. \cite{9488215} applied the decoder architecture-based I-UNN to accelerated MRI. Due to the underutilization of the MR image physic priors, I-UNN can only perform satisfactorily on randomly sampled trajectories at 4x acceleration and poorly on other common sampling scenarios such as partial Fourier and regular trajectories. In addition, \cite{heckel2020compressive} demonstrated that I-UNN exact reconstruction of the signal requires the coding matrix to satisfy the sub-Gaussian assumption, which is difficult to meet in most cases of MRI, demonstrating the theoretical limitations of I-UNN.

\subsection{$k$-Space Interpolation Methods}
Compared with the reconstruction methods in the image domain, the $k$-space interpolation methods can avoid the basis mismatches between the singularities of true continuous image and discrete grid (\cite{Ongie2016off}). Interpolating the missing data in $k$-space lies mainly in using various physical priors of MR images. Up to now, many classic works have been proposed. For example, through statistical observations or the transformation of image domain sparsity in the Fourier domain, it has been found that $k$-space data can be linearly predicted within a neighbourhood.
The most prominent example of such a method is GRAPPA (\cite{Griswold2002Generalized}). If one further assumes that the missing data can be linearly predicted by its entire neighbourhood in all channels, SPIRiT (\cite{Lustig2010spirit}) is derived.
The second one mainly capitalizes on the smooth prior of coil sensitivities. By dual relationship between the smooth prior in the image domain and the low rankness of the Hankel matrix in $k$ space, missing $k$-space prediction can be realized by a low-rank regularization problem, for which the multi-channel version of ALOHA (\cite{lee2016acceleration}) is the corresponding state-of-the-art method. The third one mainly capitalizes on phase smoothness. Like previous methods, smoothness can be transformed into low-rankness in $k$-space. The corresponding state-of-the-art method is the signal-channel version of LORAKS (\cite{6678771}). However, estimating prediction kernels or null-space filters on these methods require additional fully sampled calibration data, and the low-rankness is computationally inefficient. Therefore, it is desired to couple the physical priors characterization utilized in the above methods to the UNN for proposing a more efficient $k$-space interpolation method.

\section{Notations}\label{sect2}
In this paper, in the case of no ambiguity, matrices and vectors are all represented by bold lower case letters, i.e. $\x$, $\y$. In addition, $\x_{i}$ ($\x_{i,j}$) refers to the $i$-th column ($(i,j)$th entry) of the matrix $\x$, and $x_i$ denotes the $i$-th element of vector $\x$. Let $\mathcal{S}$ denotes a subset of indices, i.e. $\mathcal{S}\subseteq\{1,\ldots,d\}$, while $\x_{\mathcal{S}}\in \mathbb{C}^{|\mathcal{S}|}$ is the subvector of $\x\in \mathbb{C}^d$ whose entries are listed in $\mathcal{S}$. $\bm{e_k}$ denotes the $k$th standard basis vector in $\mathbb{R}^d$, equal to 1 in component $k$ and 0 everywhere else.
The superscript ${}^{H}$ for a vector or matrix denotes the Hermitian transpose. The notation $B(s)\subseteq \mathbb{C}^d$ denotes a closed boll in $\mathbb{C}^d$, i.e. $B(s)=\{\x\in\mathbb{C}^d|\|\x\|\leq s\}$. The notation $\text{supp}(f)$ for any function $f:\mathbb{C}^d\rightarrow\mathbb{C}$ denotes the support set of $f$, i.e., $\text{supp}(f):=\{\x\in\mathbb{C}^d|f(\x)\neq0\}$. The notation $\text{FFT}$ or superscript $\widehat{\cdot}$ denotes the Fourier transform.

A variety of norms on matrices will be discussed. The spectral norm of a matrix $\x$ is denoted by $\|\x\|$. The Euclidean inner product between two matrices is $\langle\x,\y\rangle=\text{Tr}(\x^H\y)$ and the corresponding Euclidean norm, termed Frobenius norm, is denoted as $\|\x\|_F$ which is derived by $\|\x\|_F:=\langle\x,\x\rangle$. The maximum entry of $\x$ is denoted as $\|\x\|_{\infty}:=\max|\x_{i,j}|$. For vectors, $\|\cdot\|$ denotes the $\ell_2$ norm. Now, let us review the matrix manipulation form for the convolutional operation. For simplicity, we only consider the 1-D case, and its extension to higher dimensions is valid (\cite{ye2018deep}). Let $\x=[x_1,\ldots,x_n]^{T}\in\mathbb{C}^n$ and $\bm{h}=[h_1,\ldots,h_d]^{T}\in\mathbb{C}^d$. A single-output convolution of the input filter $\overline{\bm{h}}$ (which is referred to the flipped version of $\bm{h}$, i.e. the indices of $\bm{h}$ are time-reversed such that $\overline{\bm{h}}[i]=\bm{h}[-i]$.) can be represented in a matrix form:
$$\x\circledast\bm{h}=\mathcal{H}(\x,d)\overline{\bm{h}}$$
where $\circledast$ denotes the convolution operator and $\mathcal{H}(\x,d)$ denotes the wrap-around Hankel matrix.

\section{Methodology and Theory }\label{sect3}
This section presents our tripled UNN for $k$-space interpolation and then analyses its interpolating accuracy guarantees under several commonly used sampling trajectories.

\subsection{The Tripled UNN}
In MRI, the forward model of multichannel $k$-space data acquisition can be expressed as
\begin{equation}\label{eq:1}\y=\mathcal{M}\widehat{\x}^*+\n\end{equation}
where $\widehat{\cdot}$ denotes an abbreviation for Fourier transform, $\widehat{\x}^*,\y,\n\in\mathbb{C}^{N\times N_c}$, $N_c\geq1$ denotes the number of channels, $\widehat{\x}^*$ denotes the true fully sampled $k$-space data, $\y$ is the under-sampled data, $\n$ is the noise and $\mathcal{M}$ denotes the sampling trajectory. In particular,
$\mathcal{M}\widehat{\x} = [M_{\Omega}\widehat{\x}_1,\ldots,M_{\Omega}\widehat{\x}_{N_c}]$, $\widehat{\x}_i$ denotes the $i$th column of $\widehat{\x}$ and $M_{\Omega}$ is a sub-sampling mask with sampling set $\Omega$, so that the $j$th diagonal of $M_{\Omega}$ is 1 if $j\in\Omega$ and zero
otherwise. The matrix $M_{\Omega}$ does not satisfy the sub-Gaussian condition.

\begin{figure}[!t]
\centering
\includegraphics[width=0.45\textwidth,height=0.28\textwidth]{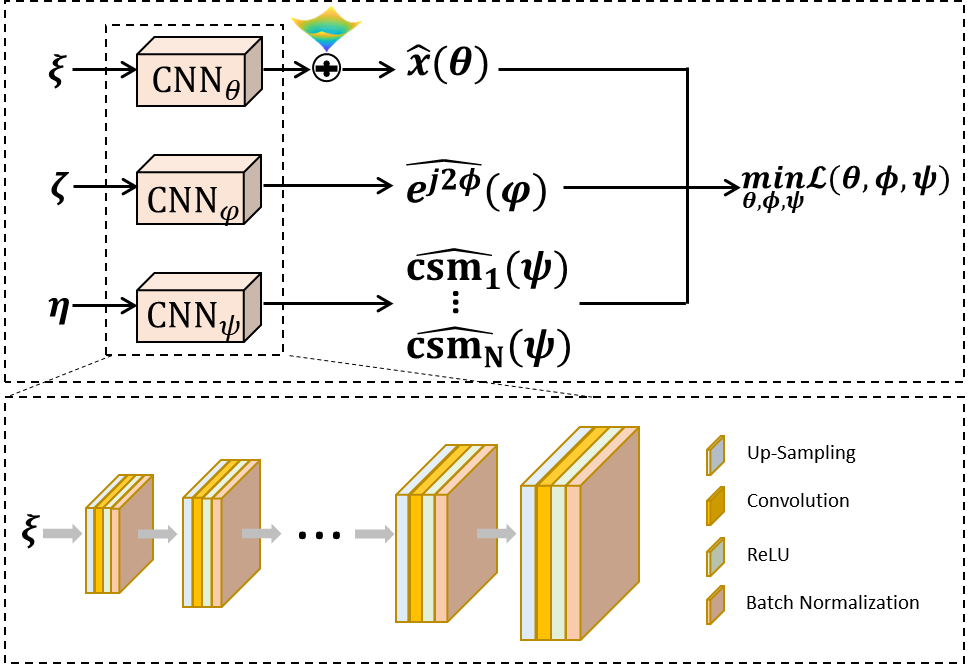}
\caption{Illustration of the proposed tripled UNN architecture. Specifically, following (\cite{8756028}), we firstly use a decoder architecture based CNN ($\widehat{\z}:=\text{CNN}_{\bm{\theta}}(\bm{\xi})$) to generalize the framelet representation drive by low-rank Hankel matrix $\mathcal{H}(\widehat{\z},d)$. Inspired by the success of (\cite{8756028}), a weighting strategy is imposed to reduce the noise amplification. We use two CNNs with small output size to represent the compact support priors, ie, $[\widehat{\text{csm}_1},\ldots,\widehat{\text{csm}_{Nc}}]:=\text{CNN}_{\bm{\varphi}}(\bm{\zeta})$, $\widehat{e^{j2\bm{\phi}}}:= \text{CNN}_{\bm{\psi}}(\bm{\eta})$, where $\bm{\xi,\zeta,\eta}$ are low-dimensional random variables. Here, ConvDecoder (\cite{9488215}) is an example of achieving the CNN modules ($\text{CNN}_{\bm{\theta}}$, $\text{CNN}_{\bm{\varphi}}$ and $\text{CNN}_{\bm{\psi}}$).}
\label{f2}
\end{figure}

Recalling MRI, the image $\x= [{\x}_1,\ldots,{\x}_{N_c}]$ is obtained by multiplying the desired image $\z$ by the coil sensitivities $\{\text{csm}_i\}$. As we know, the coil sensitivities $\{\text{csm}_i\}$ are smooth, the phase $\bm{\phi}$ is smooth, and the image $\z$ is sparse. Therefore, the MR reconstruction problem can be reduced to seeking a set of smooth coil sensitivities, a smooth phase, and a sparse image representing the multi-channel MR image. Let $\{\text{csm}_i\}$, $\bm{\phi}$ and $\z$ be variables, the MR reconstruction mathematical model reads:
\begin{equation*}\label{prior1}\left\{\begin{aligned}&\begin{aligned}
\min_{\text{csm}_i,\z,\bm{\phi}}&\sum_{i=1}^{N_c}\|\mathcal{M}\text{FFT}(\text{csm}_i\z)-\y_i\|^2\\
&+\sum_{i=1}^{N_c}\|\mathcal{M}\text{FFT}(\text{csm}_i\z^He^{j2\bm{\phi}})-\y_i\|^2\end{aligned}\\
&\begin{aligned}\text{s.t.}~ &\|\nabla\z\|_0\leq s &&(\nabla\z~ \text{is sparse})\\
 &\text{csm}_i\in \mathcal{C}^{\infty}&&(\text{csm}_i~\text{is smooth})\\
 &\bm{\phi}\in \mathcal{C}^{\infty}&&(\bm{\phi}~\text{is smooth})\end{aligned}
\end{aligned}\right.\end{equation*}
where the second term in the objective is because $\z=\z^He^{j2\phi}$, and $\z^H$ is the Hermitian transpose of $\z$. Given the dual relationship between sparseness in the image domain and low rankness in $k$-space and smoothness in the image domain and compact support in $k$-space, the above problem can be reformulated as
\begin{equation}\label{prior2}\left\{\begin{aligned}&\begin{aligned}
\min_{\text{csm},\z,\bm{\phi}}&\sum_{i=1}^{N_c}\|\mathcal{M}(\widehat{\z}\circledast\widehat{\text{csm}_i})-\y_i\|^2\\
&+\sum_{i=1}^{N_c}\|\mathcal{M}\left(\widehat{\z}^H\circledast\widehat{e^{j2\bm{\phi}}}\circledast\widehat{\text{csm}_i}\right)-\y_i\|^2\end{aligned}\\
&\begin{aligned}\text{s.t.} ~&\text{rank}(\mathcal{H}(\widehat{\z},d))\leq r &&(\mathcal{H}(\widehat{\z},d)~\text{is low-rank})\\
&|\text{supp}(\widehat{\text{csm}_i})|\leq l_1&&(\widehat{\text{csm}_i}~\text{has a compact support})\\
&|\text{supp}(\widehat{e^{j2\bm{\phi}}})|\leq l_2&& (\widehat{e^{j2\bm{\phi}}}~\text{has a compact support})\end{aligned}
\end{aligned}\right.\end{equation}
Relying on the low rankness of $\mathcal{H}(\widehat{\z},d)$, \cite{ye2018deep} showed that $\widehat{\z}$ can be represented by a deep convolutional framelet.
Following (\cite{8756028}), we used a decoder architecture based UNN to generalize the framelet representation, ie, $\widehat{\z}:=\text{CNN}_{\bm{\theta}}(\bm{\xi})$. For the compact support prior, we also used decoder architectures based UNNs with small output size to represent $\{\text{csm}_i\}$ and $\widehat{e^{j2\bm{\phi}}}$, ie, $[\widehat{\text{csm}_1},\ldots,\widehat{\text{csm}_{Nc}}]:=\text{CNN}_{\bm{\varphi}}(\bm{\zeta})$, $\widehat{e^{j2\bm{\phi}}}:= \text{CNN}_{\bm{\psi}}(\bm{\eta})$, where $ \bm{\xi,\zeta,\eta}$ are low-dimensional random variables and $ \bm{\theta,\varphi,\psi}$ are CNN networks parameters. 
Since the network modules can characterize all the above constraints, the constrained optimization problem (\ref{prior2}) can be transformed into an unconstrained optimization problem by bringing the network modules represented variables into the objective function. Further, for convenience, we rewrite it in the following compact form:
\begin{equation}\label{prior3}\min_{\bm{\theta},\bm{\varphi},\bm{\psi}}\mathcal{L}(\bm{\theta},\bm{\varphi},\bm{\psi}):=\|\mathcal{M}\mathcal{G}_{\bm{\theta},\bm{\varphi},\bm{\psi}}(\Xi)-Y\|^2_F\end{equation}
where $\Xi=[\bm{\xi,\zeta,\eta}]$, $\mathcal{G}_{\bm{\theta,\varphi,\psi}}(\Xi):=[\text{CNN}_{\bm{\theta}}(\bm{\xi})\circledast\text{CNN}_{\bm{\varphi}}(\bm{\zeta}),[\text{CNN}_{\bm{\theta}}(\bm{\xi})]^H\circledast\text{CNN}_{\bm{\psi}}(\bm{\eta})\circledast\text{CNN}_{\bm{\varphi}}(\bm{\zeta})]$ and $Y=[\y,\y]$. Specifically, $\mathcal{G}_{\bm{\theta,\varphi,\psi}}$ is the proposed tripled UNN to generate $k$-space data. The illustration of the proposed tripled UNN architecture is shown in Figure \ref{f2}.

\subsection{Theoretical Results}
In MRI, $k$-space data are usually acquired in random or deterministic sampling (including regular, partial Fourier, etc.) manners, so we need to analyze the interpolated accuracy for these two cases.
\subsubsection{Theoretical guarantee for random sampling }
Before proving our main result, we suppose that the generator derived from (\ref{prior3}) satisfies the following condition:
\begin{assump}\label{assup:1}

Let $\text{CNN}_{\bm{\theta^*}}$ be a component of the optimal ${\mathcal{G}}_{\bm{\theta^*,\varphi^*,\psi^*}}$ derived from (\ref{prior3}). For any input $\xib,\xib'\in B(s)$, the wrap around Hankel matrix of $\text{CNN}_{\bm{\theta^*}}(\xib)-\text{CNN}_{\bm{\theta^*}}(\xib')$ is rank-deficient, ie,
$$\text{rank}[\mathcal{H}(\text{CNN}_{\bm{\theta^*}}(\xib)-\text{CNN}_{\bm{\theta^*}}(\xib'),d)]\leq r,~r<d.$$
\end{assump}
\begin{rmk}
Since network $\text{CNN}_{\bm{\theta^*}}$ is Lipschitz continuous (\cite{pmlr-v70-bora17a}), if the radius $s$ of the ball $B(s)$ is small enough, the structure of the images generated by $\xib,\xib'\in B(s)$ will change little, i.e., $\widehat{\text{CNN}_{\bm{\theta^*}}(\xib)}\approx\widehat{\text{CNN}_{\bm{\theta^*}}(\xib')}$.
As a result, $\widehat{\text{CNN}_{\bm{\theta^*}}(\xib)}-\widehat{\text{CNN}_{\bm{\theta^*}}(\xib')}$ is likely to be sparse. By the Prony's result (\cite{1003065}), there is at least a non-zero filter $\bm{h}$ such that $(\text{CNN}_{\bm{\theta^*}}(\xib)-\text{CNN}_{\bm{\theta^*}}(\xib'))*\bm{h}=0$. Then, the above assumption holds.
\end{rmk}

Based on the above assumption, we have the following result:
\begin{thm}\label{thm:1}
Let ${\mathcal{G}}_{\bm{\theta^*,\varphi^*,\psi^*}}:B(s)^3\rightarrow \mathbb{R}^{N\times N_c\times2}$ be the optimal UNN derived from (\ref{prior3}).
Suppose Assumption \ref{assup:1} holds, trajectory $\mathcal{M}$ is generated independently and uniformly random with $n$ nonzero locations. For any $\x^*\in \mathbb{C}^{N\times N_c}$ and observation $\y = \mathcal{M}\x^*+\n$, there exists a constant $c_1:=\sqrt{{16n\mu_0r(N+d)\beta\log(d)}/{3N^2}}- {n}/{N}$ such that the reconstruction ${\mathcal{G}}_{\bm{\theta^*,\varphi^*,\psi^*}}(\Xi)$ satisfies:
$$\|{\mathcal{G}}(\Xi)-X^*\|_F\leq\|\widetilde{X}-X^*\|_F+\frac{2\|\mathcal{M}(\widetilde{X}-X^*)\|_F+4\|\n\|_F}{c_1} $$
with probability at least $1-2d^{2-2\beta}$ provided that $n>5.34\mu_0r(N+d)\beta\log(d)$, where ${\mathcal{G}}$ is short for ${\mathcal{G}}_{\bm{\theta^*,\varphi^*,\psi^*}}$, $X^*:=[\x^*,\x^*]$ and $\widetilde{X}\in \arg\min_{X\in {\mathcal{G}}(B(s),\bm{\zeta},\bm{\eta})}\|X-X^*\|_F$.
\end{thm}

The proof is presented in Appendix \ref{app:1}.
\begin{rmk}
\cite{YAROTSKY2017103} showed that deep neural networks can characterize the Sobolev space (including $k$-space), so it is reasonable to believe that the $\|\widetilde{X}-X^*\|_F$ term can be bounded tightly if the generator network is deep enough.
\end{rmk}
\subsubsection{Theoretical guarantee for deterministic sampling }
In practice, $k$-space data are sometimes acquired by deterministic sampling. However, to the best of our knowledge, there is currently no theoretical guarantee about the accuracy of the UNN method for reconstructing MR images or interpolating $k$-space data for deterministic sampling. Now, let us bridge this theoretical gap.
\begin{thm}\label{thm:2}Let ${\mathcal{G}}_{\bm{\theta^*,\varphi^*,\psi^*}}:B(s)^3\rightarrow \mathbb{R}^{N\times N_c\times2}$ be the optimal UNN derived from (\ref{prior3}).
Suppose ${\mathcal{G}}_{\bm{\theta^*,\varphi^*,\psi^*}}$ is bounded over $B(s)^3$, trajectory $\mathcal{M}$ is generated by deterministic sampling. For any $\x^*\in \mathbb{C}^{N\times N_c}$ and observation $\y = \mathcal{M}\x^*+\n$, there exists a constant $c_2>0$ such that the reconstruction ${\mathcal{G}}_{\bm{\theta^*,\varphi^*,\psi^*}}(\Xi)$ satisfies:
$$\|{\mathcal{G}}(\Xi)-X^*\|_F\leq\|\widetilde{X}-X^*\|_F+\frac{2\|\mathcal{M}(\widetilde{X}-X^*)\|_F+4\|\n\|_F}{c_2}$$
where ${\mathcal{G}}$ is short for ${\mathcal{G}}_{\bm{\theta^*,\varphi^*,\psi^*}}$, $X^*:=[\x^*,\x^*]$ and $\widetilde{X}\in \arg\min_{X\in {\mathcal{G}}(B(s)^3)}\|X-X^*\|_F$.
\end{thm}
The proof is presented in Appendix \ref{app:2}.
\begin{rmk}
The bounded assumption can be derived by the Lipschitz continuous of ${\mathcal{G}}_{\bm{\theta^*,\varphi^*,\psi^*}}$ (\cite{pmlr-v70-bora17a}), i.e., $\|{\mathcal{G}}_{\bm{\theta^*,\varphi^*,\psi^*}}(B(s)^3)\|\leq Ls^3$, where $L$ is the Lipschitz constant.
\end{rmk}
\section{Implementation}\label{sect4}
The evaluation was performed on the knee and brain MR data with various $k$-space trajectories, including random and deterministic (variable density regular, partial Fourier) cases. The details of the MR data are as follows:
\subsection{Data Acquisition}
\subsubsection{Knee data}
Firstly, we tested our proposed method on knee $k$-space data \footnote{\url{http://mridata.org/}}. The raw data was acquired from a 3T Siemens scanner. The number of coils was 15, and the 2D Cartesian turbo spin echo (TSE) protocol was used. The parameters for data acquisition are as follows: the repetition time (TR) was 2800ms, the echo time (TE) was 22ms, the matrix size was $768\times 770\times 1$ and the field of view (FOV) was $280 \times 280.7 \times 4.5 \text{mm}^3$. The readout oversampling was removed by transforming the $k$-space to the image and cropping the centre $384 \times 384$ region. Our proposed method does not require any additional training set. We selected data from seven subjects (including 227 slices) as the training set for trained comparison experiments.
\subsubsection{Brain data}
The raw data was acquired from a 3T Siemens scanner on a healthy female volunteer. The number of coils was 32, and the Cartesian 2D gradient echo (GRE) protocol was used. Imaging parameter included: imaging resolution = $1\times1$ mm2, TR/TE = 250/5 ms, flip angle (FA) = 70°, slice thickness = 5 mm. The maximum amplitude of wave gradient was 3.72 mT/m with cycle = 8 and readout oversampling (OS) ratio = 2, FOV $= 250 \times 250 \text{mm}^2$.
\subsubsection{Sampling trajectories}
\begin{figure}[!t]
\centering
\includegraphics[width=0.4\textwidth,height=0.15\textwidth]{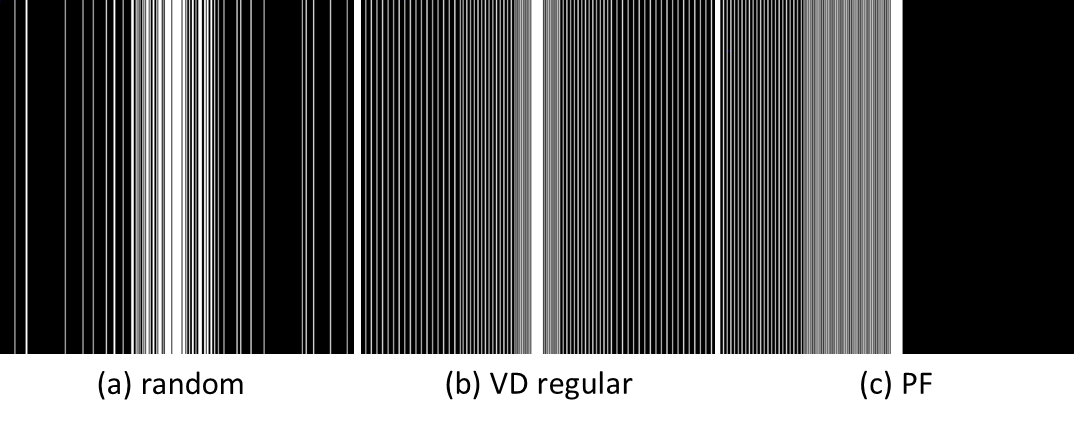}
\caption{Various sampling trajectories: (a) random undersampling at $R=5$, (b) vd regular undersampling at $R=5$, and (c) pf undersampling at $R=5$.}
\label{f4}
\end{figure}

Three undersampling trajectories were considered, including random and deterministic (variable density (VD) regular, partial Fourier (PF)) trajectories. A visualization of these sampling trajectories is depicted in Figure \ref{f4}.

\subsection{Network Architecture and Training}
The schematic diagram of the proposed tripled UNN architecture is illustrated in Figures \ref{f2}. Particularly, for CNN modules ($\text{CNN}_{\bm{\theta}}$, $\text{CNN}_{\bm{\varphi}}$ and $\text{CNN}_{\bm{\psi}}$), we used ConvDecoder\footnote{\url{https://github.com/MLI-lab/ConvDecoder}} with
\{\#layers, \#channels, \#output size\} = \{10, 256, 384$\times$384\}, \{5, 64, 11$\times$11\}, \{5, 64, 11$\times$11\} to achieve them, respectively.

For proposed tripled UNN, ADAM (\cite{kingma2014adam}) optimizer with $\beta_1=0.9, \beta_2=0.999$ is chosen for optimizing loss function (\ref{prior3}). The number of iterations is chosen as 1000. The learning rate is set to $10^{-4 }$. The models were implemented on an Ubuntu 20.04 operating system equipped with an NVIDIA A6000 Tensor Core (GPU, 48 GB memory) in the open PyTorch 1.10 framework (\cite{paszke2019pytorch}) with CUDA 11.3 and CUDNN support. For each slice, the proposed tripled UNN takes about 60s to perform iterations.

\subsection{Performance Evaluation}
In this study, the quantitative evaluations were all calculated on the image domain. The reconstructed and reference images were derived using an inverse Fourier transform followed by an elementwise square-root of sum-of-the squares (SSoS) operation, i.e. $\z[n]=(\sum_{i=1}^{N_c}|\x_i[n]|^2)^{\frac{1}{2}}$, where $\z[n]$ denotes the $n$-th element of image $\z$, and $\x_i[n]$ denotes the $n$-th element of the $i$th coil image $\x_i$. For quantitative evaluation, the peak signal-to-noise ratio (PSNR), normalized mean square error (NMSE) value, and structural similarity (SSIM) index (\cite{1284395}) were adopted.

\section{Experimentation Results}\label{sect5}
\subsection{Ablation Studies}
In order to verify that UNN in $k$-space (termed K-UNN) can more accurately characterize the physical priors of MR images compared to the traditional methods, we designed the following ablation experiments. First, for the multichannel parallel imaging, the phase module $\text{CNN}_{\bm{\psi}}(\bm{\eta})$ was removed from the tripled K-UNN (\ref{prior3}) and compared with the L1-SPIRiT (\cite{Lustig2010spirit}) to verify the accuracy of K-UNN for the characterization of sensitivity prior. Then, for single channel partial Fourier imaging, the coil sensitivity module $\text{CNN}_{\bm{\varphi}}(\bm{\zeta})$ was removed from the tripled K-UNN (\ref{prior3}) and compared with the data fitting, and $k$-space convolution (K-COV) method (\cite{huang2009partial}) to verify the accuracy of K-UNN for the characterization of phase prior. Finally, for multichannel parallel imaging, the L1-SENSE-LORAKS (\cite{kim2017loraks}) (for which a wavelet-based L1 penalty is added on its publicly available Matlab code \footnote{\url{https://mr.usc.edu/download/LORAKS2/}}) was compared to verify the joint characterization ability of the proposed tripled K-UNN for the sparse, coil sensitivity smooth, and phase smooth priors. Our code is available at \footnote{\url{https://github.com/ZhuoxuCui/K_UNN}}.
\subsubsection{Characterization of sensitivity prior}
\begin{figure*}[!t]
\centering
\includegraphics[width=0.96\textwidth,height=0.36\textwidth]{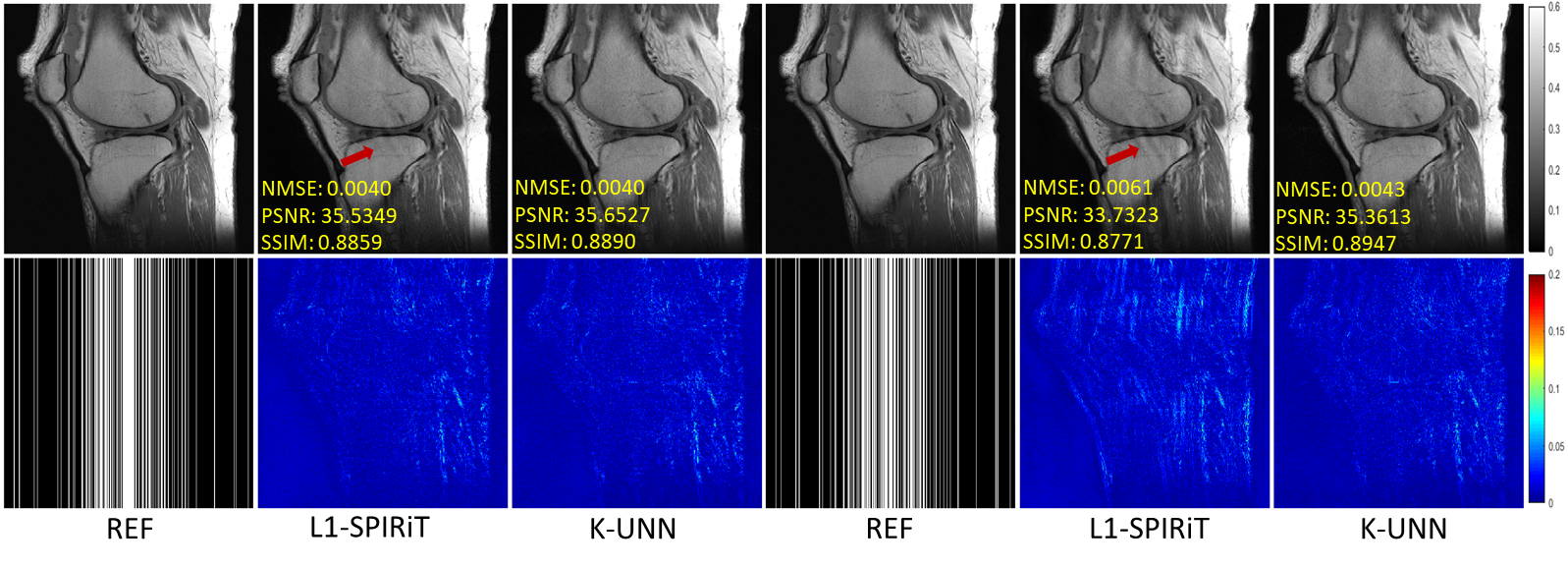}
\caption{The left part illustrates the reconstruction results of 16 ACS lines at $R=4$, and the right part illustrates the reconstruction results of 6 ACS lines at $R=4$. The values in the corner are NMSE/PSNR/SSIM values of each slice. The second row illustrates the error views. The grayscale of the reconstructed images and the color bar of the error images are at the right of the figure.}
\label{fa1}
\end{figure*}

In this section, we eliminate the effect of phase and verify the ability of K-UNN to characterize the coil sensitivity prior. In particular, by eliminating the second term in the objective function (\ref{prior2}), the $k$-space data is generated only by $\text{CNN}_{\bm{\theta}}(\bm{\xi})\circledast\text{CNN}_{\bm{\varphi}}(\bm{\zeta})$. As shown in Figure \ref{fa1}, when ACS data is abundant (16-lines), both our K-UNN and L1-SPIRiT can reconstruct the image satisfactorily. However, a closer look shows that in the region of interest of the knee cartilage, a few artifacts remain in the L1-SPIRiT reconstruction. When the ACS data is insufficient (6-lines), the L1-SPIRiT reconstruction exhibits artifacts, while the quality of the reconstructed image by our K-UNN deteriorates negligibly. Therefore, the above experimental results verify the superiority of the proposed K-UNN in the characterization of coil sensitivity prior.

\subsubsection{Characterization of phase prior}
\begin{figure*}[!t]
\centering
\includegraphics[width=0.96\textwidth,height=0.6\textwidth]{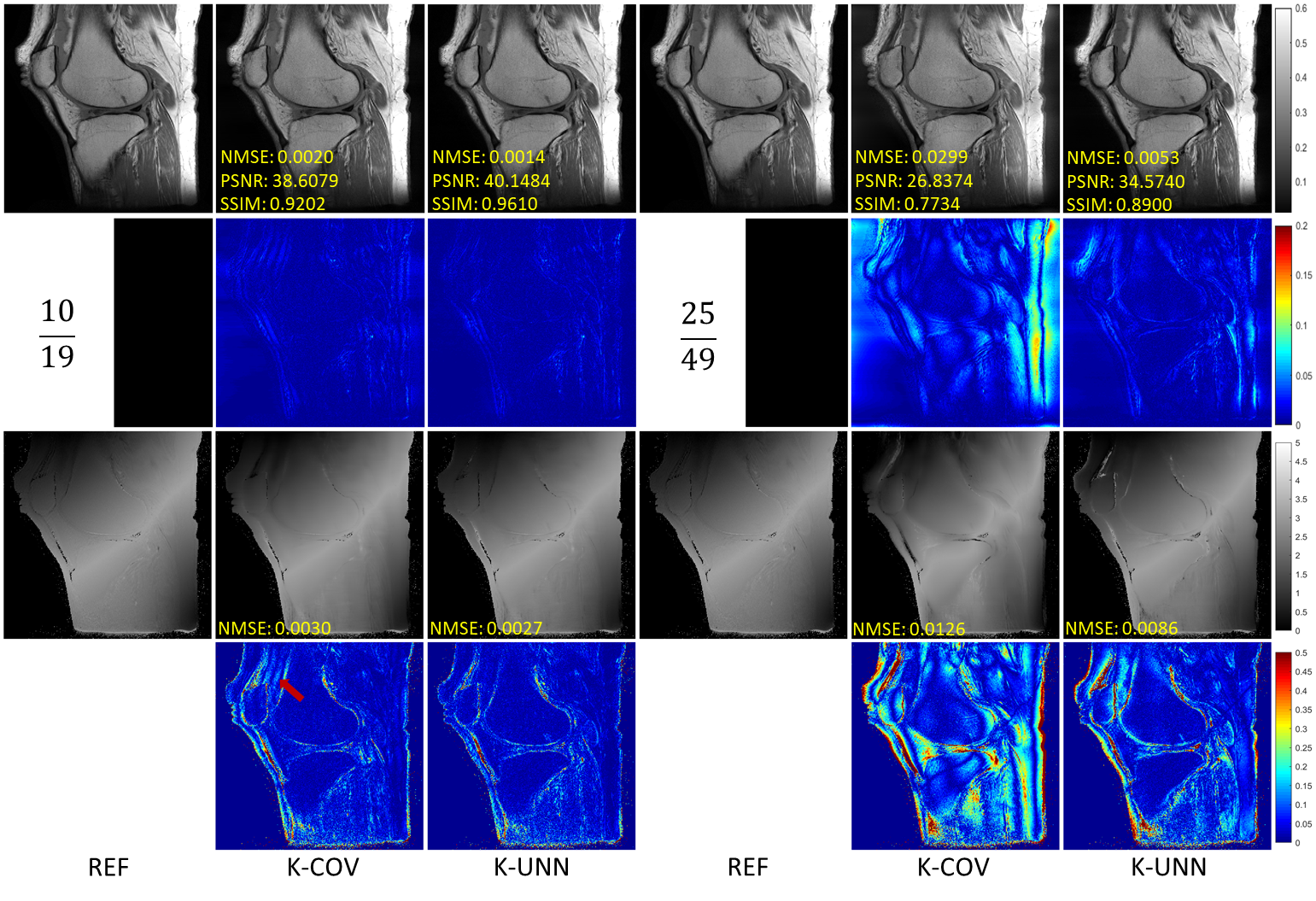}
c\caption{The left part illustrates the reconstruction results at $R=19/10$, and the right part illustrates the reconstruction results at $R=49/25$. The values in the corner are NMSE/PSNR/SSIM values of each slice. The first and third rows illustrate the reconstructed images and their phases, respectively. The second and fourth rows illustrate the error views of reconstructed images and phases, respectively. The grayscale and color bar are at the right of the figure.}
\label{fa2}
\end{figure*}

In this section, we eliminate the effect of coil sensitivities and verify the ability of K-UNN to characterize the phase prior. Specifically, we first merged the knee data into a single channel. For single-channel partial Fourier imaging, the ablation experiments can be performed directly by removing the coil sensitivity module $\text{CNN}_{\bm{\varphi}}(\bm{\zeta})$ in the model (\ref{prior3}). As shown in Figure \ref{fa2}, both the proposed K-UNN and K-COV can reconstruct the image satisfactorily when ACS data is abundant ($R=19/10$). However, it can be seen from the phase error that K-COV biases the phase reconstruction in the interior of the image (indicated by the red arrow). When the ACS data is insufficient ($R=49/25$), it is clear that our method outperforms K-COV in terms of both image reconstruction quality and phase reconstruction quality. This experiment verifies the proposed K-UNN's superiority in the prior characterization of the phase.

\subsubsection{Joint characterization of physical priors}
\begin{figure*}[!t]
\centering
\includegraphics[width=0.96\textwidth,height=0.32\textwidth]{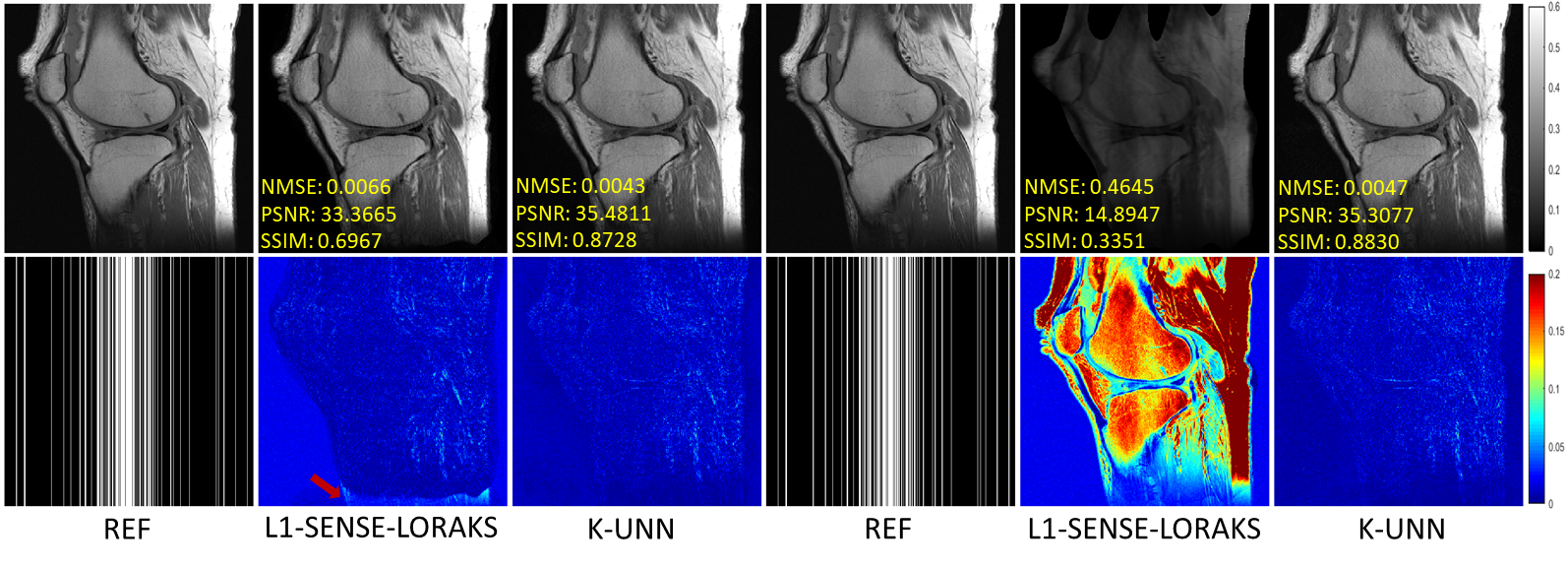}
\caption{The left part illustrates the reconstruction results of 10 ACS lines at $R=5$, and the right part illustrates the reconstruction results of 8 ACS lines at $R=5$. The values in the corner are NMSE/PSNR/SSIM values of each slice. The second row illustrates the error views. The grayscale of the reconstructed images and the color bar of the error images are at the right of the figure.}
\label{fa3}
\end{figure*}

In this section, we verify the joint characterization ability of the proposed K-UNN for the sparse, coil sensitivity smooth, and phase smooth priors. As shown in Figure \ref{fa3}, when ACS data is abundant (10-lines), our K-UNN and L1-SENSE-LORAKS can satisfactorily reconstruct the image. When the ACS data are insufficient (8-lines), the reconstruction quality of L1-SENSE-LORAKS drops sharply due to inaccurate coil sensitivities estimation, while the quality of the reconstructed image by our K-UNN only deteriorates slightly. This experiment verifies the superiority of the proposed K-UNN in the joint characterization of three priors.

\subsection{Comparative Studies}
In this section, to demonstrate the effectiveness of our K-UNN, a series of extensive comparative experiments were studied. In particular, we compared the traditional $k$-space PI method (L1-SPIRiT (\cite{Lustig2010spirit}), L1-SENSE-LORAKS (\cite{kim2017loraks})) and  I-UNN (\cite{9488215}) which dose not utilize physical priors. To further verify the superiority of the proposed method, we also compared it to the SOTA supervised-trained $k$-space DL method H-DSLR (\cite{9159672}), for which we develop a PyTorch-based implementation based on its publicly available TensorFlow code\footnote{\url{https://github.com/anikpram/Deep-SLR}}.

\subsubsection{Experiments on random sampling trajectory}
\begin{figure*}[!t]
\centering
\includegraphics[width=0.96\textwidth,height=0.5\textwidth]{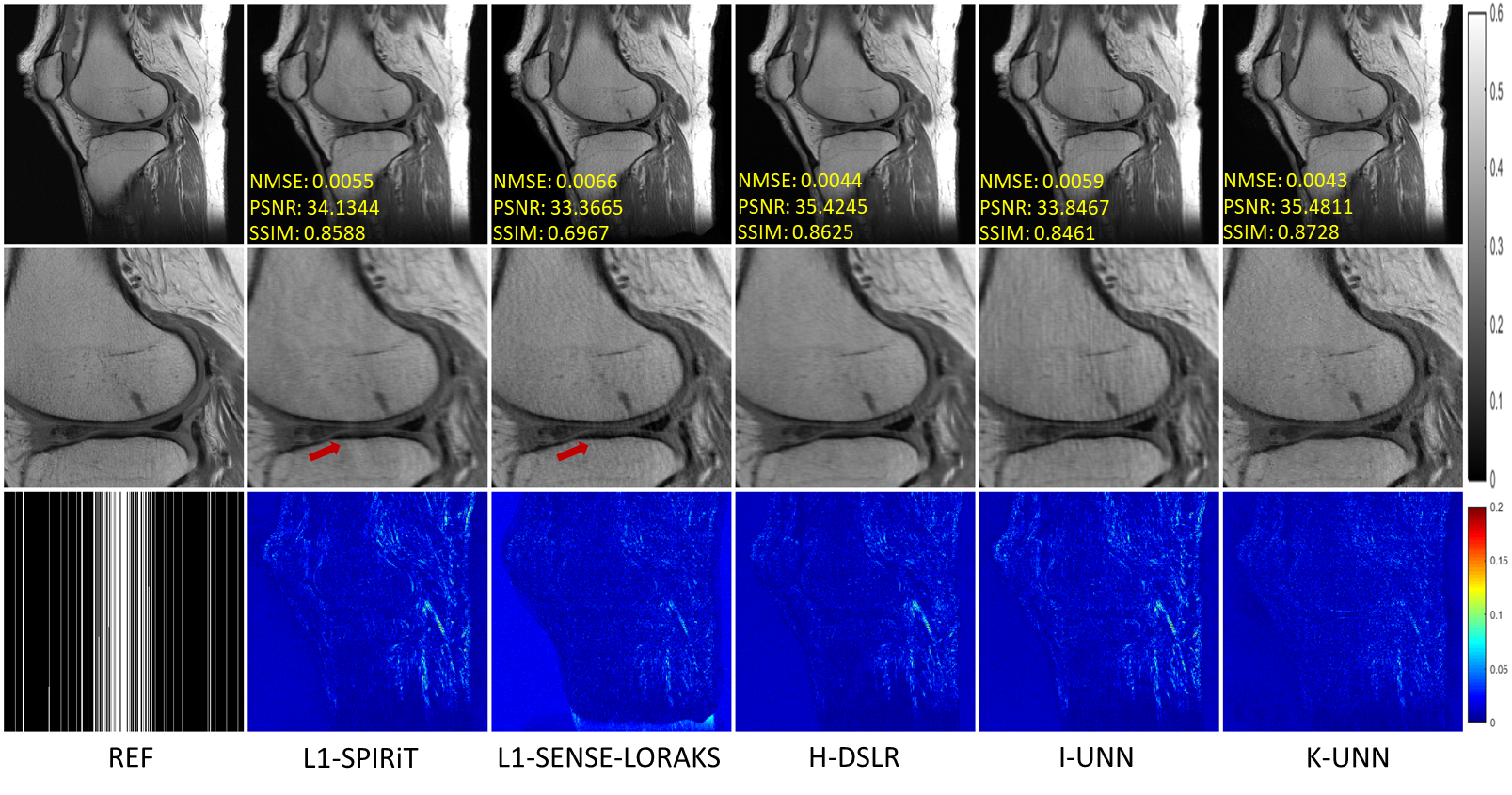}
\caption{Reconstruction results under random undersampling at $R=5$. The values in the corner are NMSE/PSNR/SSIM values of each slice. Second and third rows illustrate the enlarged and error views, respectively. The grayscale of the reconstructed images and the color bar of the error images are at the right of the figure.}
\label{f5}
\end{figure*}

In this section, we test the performances of the proposed K-UNN and comparison methods under random sampling. Figure \ref{f5} shows the reconstruction results of the knee data using various methods in the case of random sampling with an acceleration factor of 5. For L1-SPIRiT, the aliasing pattern remains in the reconstructed images. For L1-SENSE-LORAKS, there are still artifacts in the region of interest of the reconstructed image, as seen in the enlarged view. For I-UNN, although it has been empirically noted in the literature (\cite{Lustig2007Sparse}) that the random sampling composite Fourier encoding can approximate the sub-Gaussian condition, however, the acceleration exceeds its limit on this experiment, so there are still artifacts in the reconstruction images. For the SOTA trained $k$-space DL method (H-DSLR), the noise in the reconstruction image is effectively suppressed. However, the reconstructed image has a serious loss of high-frequency details in the upper right of the enlarged view. It is not difficult to find that our K-UNN can effectively suppress artifacts and recover the high-frequency detail better.

The competitive quantitative results of the above methods are shown in Table \ref{tab:1}. Our method consistently outperforms traditional methods L1-SPIRiT, L1-SENSE-LORAKS, image domain I-UNN, and SOTA-trained H-DSLR as characterized by visual and quantitative evaluations. The above experiments confirm the competitiveness of our method under a random sampling trajectory.

\begin{table}
	\begin{center}
		\caption{Quantitative comparison for various methods on the knee data under random, VD regular and PF trajectories with $R=5$.}\label{tab:1}
		\scriptsize{
			\setlength{\tabcolsep}{1.8mm}{
				\begin{tabular}{l|l|ccc}
					\hline
					\multicolumn{ 2}{c}{ Datasets} & \multicolumn{ 3}{|c}{Quantitative Evaluation}  \\
					\multicolumn{ 2}{c|}{ \& Methods   } &NMSE &PSNR(dB)&SSIM   \\
					\hline
					\multirow{6}{*}{Random ($R=5$)  }
					& L1-SPIRiT  &0.0055&34.1344&0.8588 \\
					\cline{2-5}
					& L1-SENSE-LORAKS  &0.0066&33.3665&0.6967 \\
					\cline{2-5}
					& H-DSLR &0.0044&35.4245&0.8625\\
					\cline{2-5}
					&  I-UNN &0.0059&33.8467&0.8461\\
					\cline{2-5}
					&  K-UNN &\textcolor{red}{0.0043}&\textcolor{red}{35.4811}&\textcolor{red}{0.8728}\\
					\hline
					\multirow{6}{*}{VD Regular ($R=5$)}
					& L1-SPIRiT  &0.0033&36.3754&0.8940 \\
					\cline{2-5}
					& L1-SENSE-LORAKS  &0.0049&34.6815&0.7260 \\
					\cline{2-5}
					& H-DSLR &\textcolor{red}{0.0026}&\textcolor{red}{37.3599}&\textcolor{red}{0.9045}\\
					\cline{2-5}
					&  I-UNN &0.0065&33.4455&0.8724\\
					\cline{2-5}
					& K-UNN  &0.0034&36.3166&0.9005\\
					\hline
					\multirow{6}{*}{PF ($R=5$)}
					& L1-SPIRiT  &0.0138&30.1501&0.8582\\
                    \cline{2-5}
					& L1-SENSE-LORAKS  &0.0047&34.8409&0.7558 \\
					\cline{2-5}
					& H-DSLR &0.0059&33.8551&0.8976\\
					\cline{2-5}
					&  I-UNN &0.0052&34.3695&0.8783\\
					\cline{2-5}
					&  K-UNN &\textcolor{red}{0.0037}&\textcolor{red}{35.9568}&\textcolor{red}{0.9015}\\
					\hline
		\end{tabular}}}
	\end{center}
\end{table}
\subsubsection{Experiments on deterministic sampling trajectories}
\begin{figure*}[!t]
\centering
\includegraphics[width=0.96\textwidth,height=0.5\textwidth]{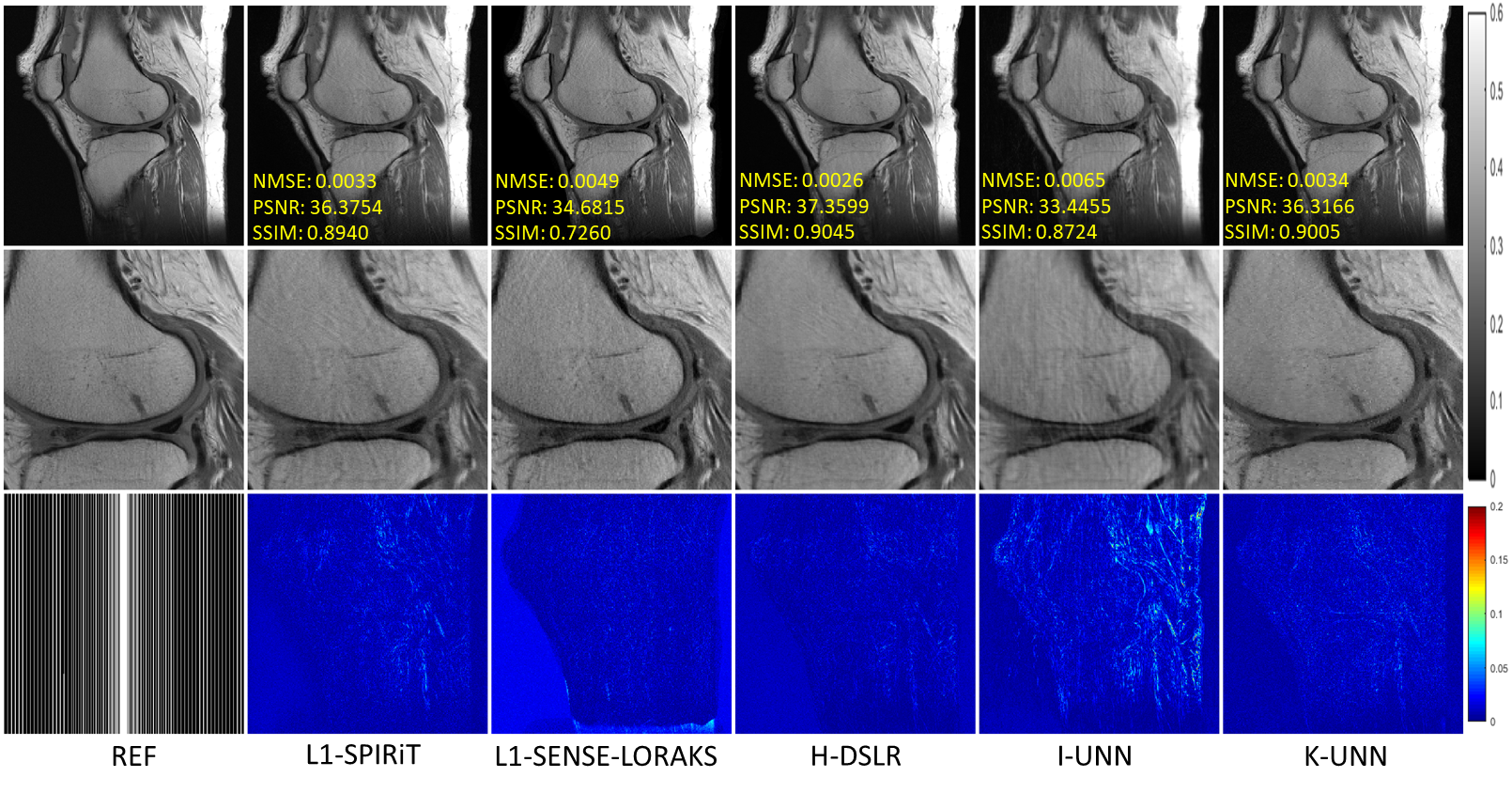}
\caption{Reconstruction results under VD regular undersampling at $R=5$. The values in the corner are NMSE/PSNR/SSIM values of each slice. Second and third rows illustrate the enlarged and error views, respectively. The grayscale of the reconstructed images and the color bar of the error images are at the right of the figure.}
\label{f6}
\end{figure*}

In this section, we test the performances of the proposed K-UNN and comparison methods under deterministic sampling trajectories, including VD regular and PF trajectories. Figure \ref{f6} shows the reconstruction results of the knee data using various methods in the case of VD regular sampling with an acceleration factor of 5. As shown in Figure \ref{f6}, for L1-SPIRiT and L1-SENSE-LORAKS, the aliasing pattern remains in the reconstructed images. Due to the serious violation of sub-Gaussian assumption by VD regular sampling composite Fourier encoding, serious artifacts remain in the I-UNN reconstructed images, which also verifies the theoretical validity reversely. Although the proposed K-UNN is slightly inferior to H-DSLR in terms of quantitative metrics, visually, the proposed method achieves better performance in artifact suppression. In this experiment, the proposed untrained method achieved comparable performance to the SOTA trained method, which is a good indication of the superiority of the proposed method.

\begin{figure*}[!t]
\centering
\includegraphics[width=0.96\textwidth,height=0.5\textwidth]{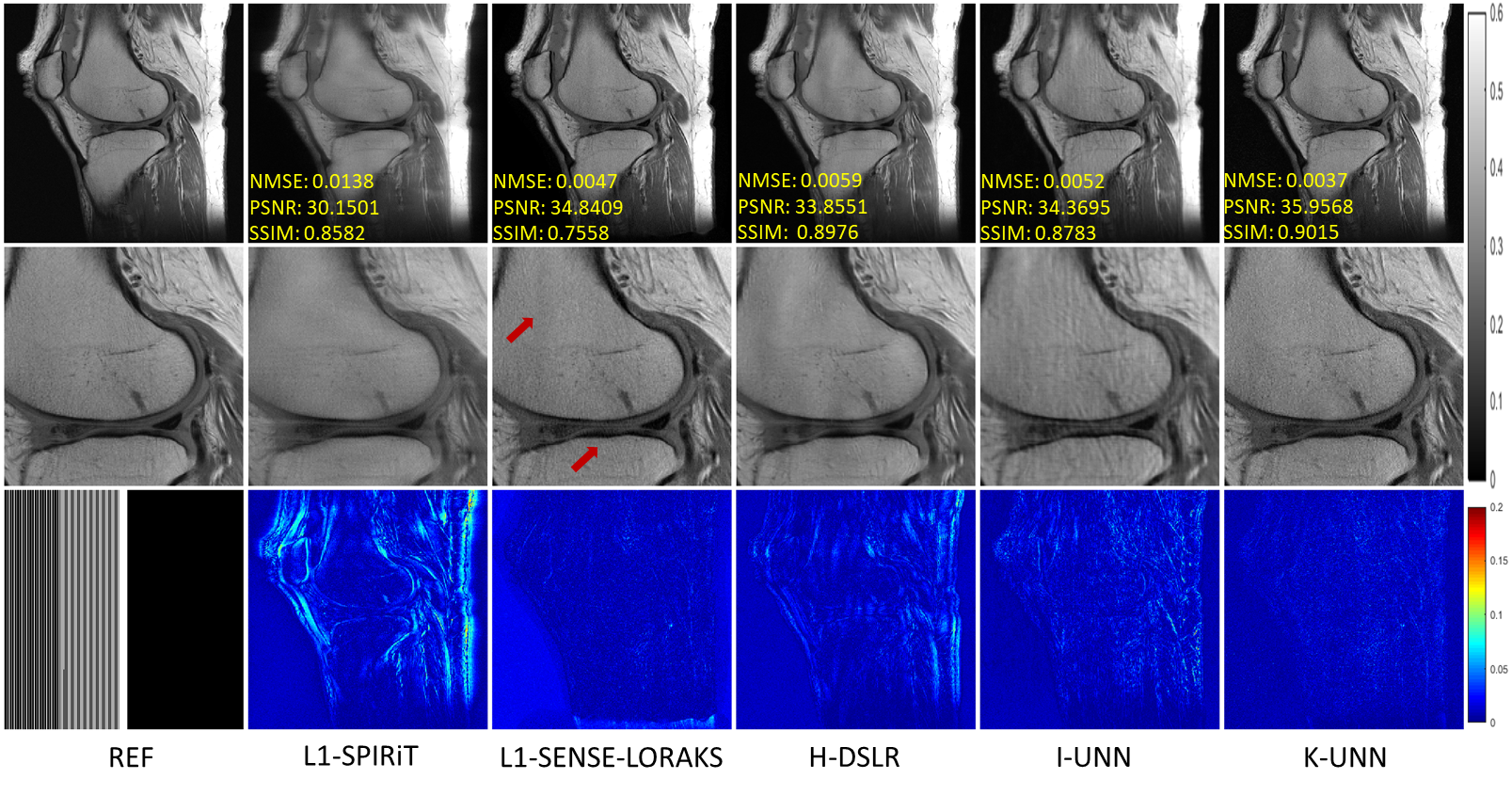}
\caption{Reconstruction results under PF undersampling at $R=5$. The values in the corner are NMSE/PSNR/SSIM values of each slice. Second and third rows illustrate the enlarged and error views, respectively. The grayscale of the reconstructed images and the color bar of the error images are at the right of the figure.}
\label{f7}
\end{figure*}

Figure \ref{f7} shows the reconstruction results of the knee data using various methods in the case of PF sampling with an acceleration factor of 5. As shown in Figure \ref{f7}, due to the absence or inability to fully utilize the phase prior, L1-SPIRiT and H-DSLR reconstructed images are blurred.
Due to the serious violation of sub-Gaussian assumption by PF sampling composite Fourier encoding, I-UNN reconstructed image exhibits artifacts. Although L1-SENSE-LORAKS can reconstruct the image well, a closer look reveals two inconspicuous artifacts at the arrow's point. Competitive quantitative results are shown in Table \ref{tab:1}. Our proposed method achieves the best performance compared to all methods in terms of quantitative metrics. Combining visual perception and quantitative indicators, we verify the effectiveness of the proposed K-UNN method under the PF sampling.

The above experiments verify the superiority of the proposed method under both random and deterministic sampling trajectories, and the correctness of our Theories \ref{thm:1} and \ref{thm:2} is further confirmed.

\section{Discussion}\label{sect6}
In this study, we propose an MR physical priors driven $k$-interpolation method for MRI using UNN, dubbed K-UNN. In theory, we analyze its interpolated accuracy bound, which has also been verified by some comparative experiments on MR image reconstruction qualities. However, some areas still need further discussion or improvement for our proposed model.

\subsection{Priors Mismatches}
\begin{figure}[!t]
\centering
\includegraphics[width=0.38\textwidth,height=0.25\textwidth]{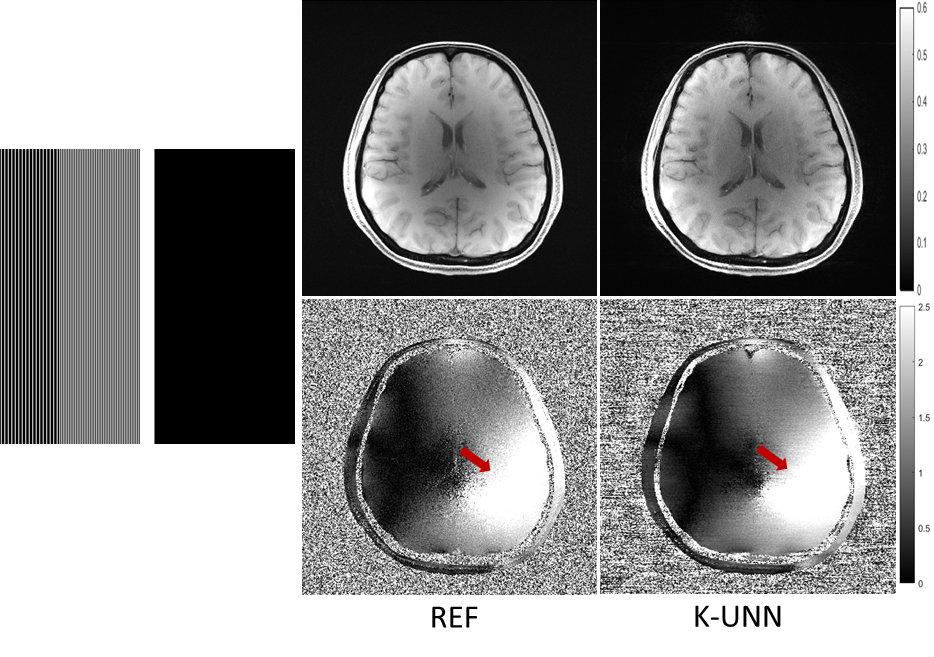}
\caption{Reconstruction results of GRE protocol acquired data under PF undersampling at $R=5$. The second row illustrates the phases of reconstructed images. The grayscale of the reconstructed images is at the right of the figure.}
\label{f8}
\end{figure}

If, in reality, there are unsatisfiable cases of image sparsity, phase smoothness, and coil sensitivity smoothness in the proposed method, the imaging model (\ref{prior2}) or (\ref{prior3}) can be recomposed flexibly from the modules as adopted in the ablation experiments. It is worth mentioning that in our model, image sparsity prior, phase and coil sensitivity smooth prior have been generalized to characterize the $k$-space learnable framelet and small size convolution kernel, respectively. The proposed model not only characterizes the priors more accurately (as has been verified by ablation experiments) but also has a wider range of applicability. For example, the phase of the image acquired by the GRE sequence is usually not particularly smooth ($\notin \mathcal{C}^{\infty}$). Figure \ref{f8} shows the reconstruction result of the GRE acquired data using K-UNN in the case of PF sampling with an acceleration factor of 5. Not only is the image accurately reconstructed, but the jump boundaries indicated by the red arrows are also accurately reconstructed.

\subsection{Further Improvements}
In the proposed method, we do not explicitly utilize the ACS data. However, in some applications of MRI, ACS data can be acquired by additional fast scan sequences. Our future work will be on how to characterize the prior from the ACS data and couple it with the proposed model.

The proposed model only makes full use of the priors of a single image or a single $k$-space data, while in practice, MRI can draw on correlation with other images. For example, in dynamic imaging, the current frame image can be regarded as the image of the previous frame that has been deformed. Therefore, our future work will propose an untrained deformation operator to capitalize inter-frame similarity priors and couple them into the proposed K-UNN.

At last, for supervised deep learning methods, their training process is offline, and the test (reconstruction) process takes very little time.
Our methods need to solve the optimization problem (\ref{prior3}) individually for each slice (task), which severely limits the speed of imaging.
Instead of optimizing (\ref{prior3}) individually, it will be a potential research direction to shorten the imaging time of our method via meta-learning the past experiences by exploiting some similarities across the previous tasks (\cite{Vilalta2002a}).

\section{Conclusion}\label{sect7}
This paper proposed a $k$-interpolation method for MRI using UNN, dubbed K-UNN. Unlike existing UNNs, the proposed K-UNN can capitalize three physical priors of the MR image (or $k$-space data), including sparsity, coil sensitivities, and phase smoothness. Theoretically, we proved that the proposed method guarantees tight bounds for interpolated $k$-space data accuracy on a series of commonly used sampling trajectories. Experiments showed that the proposed method consistently outperforms traditional parallel imaging methods and I-UNN, and even outperforms the SOTA supervised-trained $k$-space deep learning methods in some cases. Our method could be a powerful framework for parallel MR imaging, and the further development of this kind of method may enable even larger gains in the future.

\section*{Appendix}
\subsection{Proof of Theorem \ref{thm:1}}\label{app:1}
Before proving in detail, let's review some useful technical lemmas in matrix completion (\cite{candes2009exact,6867345}). Recall that any $N\times d$ rank-$r$ matrix $\bm{H}$ can be decomposed as follows via SVD decomposition:
$$ \bm{H}=\sum_{k=1}^r\sigma_k\uu_k\vv_k^H.$$
According to the left and right eigenvectors $\uu_k$ and $\vv_k$, one can define the following subspace:
\begin{equation}\label{sub}\left\{\begin{aligned}
U :=& \text{span}\{\uu_k|k=1,\ldots,r,\}\subseteq\mathbb{C}^{N} \\
V :=& \text{span}\{\vv_k|k=1,\ldots,r,\}\subseteq\mathbb{C}^{d}\\
T:=&\text{span}\{\uu_k\y^H+\x\vv_k^H|k=1,\ldots,r,\forall \y,\x\}\subseteq\mathbb{C}^{N\times d}
\end{aligned}\right.\end{equation}
The coherence between above subspaces and standard basis can be measured by the following definition:
\begin{defn}Let $U(V)$ be a subspace of dimension $r$ and ${P}_{U}$ denotes the orthogonal projection onto it, the coherence of $U$ is defined as
$$\mu(U):=\frac{n}{r}\max_{1\leq i \leq n}\|{P}_{U}\e_i\|^2$$
where $\e_i$ is the standard basis.
\end{defn}

Now, we will give a pivotal lemma for our method.
By the architecture of the optimal UNN  ${\mathcal{G}}_{\bm{\theta^*,\varphi^*,\psi^*}}(\Xi)$ derived from (\ref{prior3}), we have
\begin{equation*}\begin{aligned}
&\mathcal{G}_{\bm{\theta^*,\varphi^*,\psi^*}}(\Xi)\\
=&[\text{CNN}_{\bm{\theta}^*}(\bm{\xi})\circledast\text{CNN}_{\bm{\varphi}^*}(\bm{\zeta}),\\
&~[\text{CNN}_{\bm{\theta}^*}(\bm{\xi})]^H\circledast\text{CNN}_{\bm{\psi}^*}(\bm{\eta})\circledast\text{CNN}_{\bm{\varphi}^*}(\bm{\zeta})]\\
=&[\mathcal{H}(\text{CNN}_{\bm{\theta}^*}(\bm{\xi}),d)\overline{\text{CNN}_{\bm{\varphi}^*}(\bm{\zeta})},\\
&~\mathcal{H}([\text{CNN}_{\bm{\theta}^*}(\bm{\xi})]^H,d)\overline{\text{CNN}_{\bm{\psi}^*}(\bm{\eta})\circledast\text{CNN}_{\bm{\varphi}^*}(\bm{\zeta})}]\\
\end{aligned}\end{equation*}
Let $\mathcal{G}$, $\bm{k_1}$ and $\bm{k_2}$ be abbreviations for $\mathcal{G}_{\bm{\theta^*,\varphi^*,\psi^*}}$, $\overline{\text{CNN}_{\bm{\varphi}^*}(\bm{\zeta})}$ and $\overline{\text{CNN}_{\bm{\psi}^*}(\bm{\eta})\circledast\text{CNN}_{\bm{\varphi}^*}(\bm{\zeta})}$, respectively,
$k$-space data acquisition (\ref{eq:1}) can be reformulated as
\begin{equation}\begin{aligned}\label{matirx}&\mathcal{ M}{\mathcal{G}}(\Xi)\\
=&[M_{\Omega}\mathcal{H}(\text{CNN}_{\bm{\theta}^*}(\bm{\xi}),d)\bm{k_1},M_{\Omega}\mathcal{H}([\text{CNN}_{\bm{\theta}^*}(\bm{\xi})]^H,d)\bm{k_2}].\end{aligned}\end{equation}

Based on the above notations, we have:
\begin{lem}\label{lem:1}
Let ${\mathcal{G}}:B(s)^3\rightarrow \mathbb{R}^{N\times N_c\times2}$ be the optimal UNN derived from (\ref{prior3}).
Suppose Assumption \ref{assup:1} holds and trajectory $\mathcal{M}$ is generated independently and uniformly with $n$ nonzero locations. For any $\Xi,\Xi'\in B(s)^3$, $\Xi:=(\xib,\bm{\zeta},\bm{\eta})$, and $\Xi':=(\xib',\bm{\zeta},\bm{\eta})$, we have
\begin{equation}\label{ineq:1}\|\mathcal{M}{\mathcal{G}}(\Xi)-\mathcal{M}{\mathcal{G}}(\Xi')\|_F\geq c \|{\mathcal{G}}(\Xi)-{\mathcal{G}}(\Xi')\|_F\end{equation}
with probability at least $1-2d^{2-2\beta}$ provided that $n>16/3\mu_0r(N+d)\beta\log(d)$, where subspaces $U$, $V$ and $T$ (see (\ref{sub})) are spanned by the right and left singular vectors of
$\mathcal{H}(\text{CNN}_{\bm{\theta}^*}(\xib)-\text{CNN}_{\bm{\theta}^*}(\xib'),d)$ and $\mu_0$ is a constant satisfying $\mu_0\geq\max\{\mu(U),\mu(V)\}$.
\end{lem}

By Assumption \ref{assup:1}, we have $\text{rank}[\mathcal{H}(\text{CNN}_{\bm{\theta}^*}(\xib)-\text{CNN}_{\bm{\theta}^*}(\xib'),d)]\leq r$ with $r<\min\{N,d\}$. Following Theorem 6 of literature (\cite{Recht2011a}), we know that the following inequality holds with a probability at least $1-2n_2^{2-2\beta}$ and $\beta>1$ for the subspace $T$ spanned by the right and left singular vectors of
$\mathcal{H}(\text{CNN}_{\bm{\theta}^*}(\xib)-\text{CNN}_{\bm{\theta}^*}(\xib'),d)$:
\begin{equation*}
\frac{N}{n}\left\|\mathcal{P}_T\mathcal{R}_{\Omega}\mathcal{P}_T-\frac{n}{N}\mathcal{P}_T\right\|\leq\sqrt{\frac{16\mu_0r(N+d)\beta\log(d)}{3n}}
\end{equation*}
where $\mathcal{R}_{\Omega}(\bm{H}):=\sum_{a\in\Omega}\sum_{b=1}^d\langle \bm{e}_a\bm{e}_b^H,\bm{H}\rangle\bm{e}_a\bm{e}_b^H$, which meets $\mathcal{R}_{\Omega}(\bm{H})=M_{\Omega}\bm{H}$.
The above inequality implies
$$\mathcal{P}_T\mathcal{R}_{\Omega}\mathcal{P}_T\succeq \sqrt{\frac{16n\mu_0r(N+d)\beta\log(d)}{3N^2}}\mathcal{I}- \frac{n}{N}\mathcal{P}_T.$$
Since $\mathcal{H}(\text{CNN}_{\bm{\theta}^*}(\xib)-\text{CNN}_{\bm{\theta}^*}(\xib'),d)\in T$ and $\mathcal{H}(\text{CNN}_{\bm{\theta}^*}(\xib)-\text{CNN}_{\bm{\theta}^*}(\xib'),d)=\mathcal{H}(\text{CNN}_{\bm{\theta}^*}(\xib),d)-\mathcal{H}(\text{CNN}_{\bm{\theta}^*}(\xib'),d)$, we have
 \begin{equation*}\begin{aligned}
&\|M_{\Omega}\mathcal{H}(\text{CNN}_{\bm{\theta}^*}(\xib),d)\bm{k_1}-M_{\Omega}\mathcal{H}(\text{CNN}_{\bm{\theta}^*}(\xib'),d)\bm{k_1}\|_F\\
\geq&\|\mathcal{P}_T\mathcal{R}_{\Omega}\mathcal{P}_TM_{\Omega}\mathcal{H}(\text{CNN}_{\bm{\theta}^*}(\xib),d)\bm{k_1}\\
&-\mathcal{P}_T\mathcal{R}_{\Omega}\mathcal{P}_T\mathcal{H}(\text{CNN}_{\bm{\theta}^*}(\xib'),d)\bm{k_1}\|_F\\
\geq&c\|\mathcal{P}_T\mathcal{H}(\text{CNN}_{\bm{\theta}^*}(\xib),d)\bm{k_1}-\mathcal{P}_T\mathcal{H}(\text{CNN}_{\bm{\theta}^*}(\xib'),d)\bm{k_1} \|_F\\
=&c\|\mathcal{H}(\text{CNN}_{\bm{\theta}^*}(\xib),d)\bm{k_1}-\mathcal{H}(\text{CNN}_{\bm{\theta}^*}(\xib'),d)\bm{k_1} \|_F\\
\end{aligned}\end{equation*}
with $c:=\sqrt{{16n\mu_0r(N+d)\beta\log(d)}/{3N^2}}- {n}/{N}$. It is not difficult to verify that the above inequality also holds for $M_{\Omega}\mathcal{H}([\text{CNN}_{\bm{\theta}^*}(\bm{\xi})]^H,d)\bm{k_2}-M_{\Omega}\mathcal{H}([\text{CNN}_{\bm{\theta}^*}(\bm{\xi'})]^H,d)\bm{k_2}$
Then, the inequality (\ref{ineq:1}) follows by applying the equation (\ref{matirx}) directly.

Now, let us start to give the concrete proof for Theorem \ref{thm:1}.

Firstly, we define the following two subsets $\mathcal{C}_1$ and $\mathcal{C}_2$ such that
\begin{equation*}\begin{aligned}
&\Xi=(\xib,\bm{\zeta},\bm{\eta})\in\mathcal{C}_1=\arg\min_{(\xib,\bm{\zeta},\bm{\eta})\in B(r)^3}\|Y-\mathcal{M}{\mathcal{G}}((\xib,\bm{\zeta},\bm{\eta}))\|_F\\
&\Xi'=(\xib',\bm{\zeta},\bm{\eta})\in\mathcal{C}_2=\arg\min_{\xib'\in B(r)}\|X^*-{\mathcal{G}}((\xib',\bm{\zeta},\bm{\eta}))\|_F\\
\end{aligned}\end{equation*}
Using the inequality (\ref{ineq:1}) directly, we have
 \begin{equation*}\begin{aligned}
\|{G}(\Xi)-{G}(\Xi')\|_F\leq&\frac{1}{c}\|\mathcal{M}{G}(\Xi)-\mathcal{M}{G}(\Xi')\|_F\\
\leq&\frac{1}{c}\|\mathcal{M}{G}(\Xi)-Y\|_F+\frac{1}{c}\|\mathcal{M}{G}(\Xi')-Y\|_F\\
\leq&\frac{2}{c}(\|\mathcal{M}({G}(\Xi')-X^*)\|_F+2\|\n\|_F)\\
\end{aligned}\end{equation*}
where we make use of the notation $\|\mathcal{M}{G}(\Xi')-Y\|_F\leq \|\mathcal{M}{G}(\Xi)-Y\|_F$ for the last inequality.
Then it holds that
\begin{equation*}\begin{aligned}
&\|{G}(\Xi)-X^*\|_F\\
\leq&\|{G}(\Xi')-X^*\|_F+\|{G}(\Xi)-{G}(\Xi')\|_F\\
\leq&\|{G}(\Xi')-X^*\|_F+\frac{2\|\mathcal{M}({G}(\Xi')-X^*)\|_F+4\|\n\|_F}{c}\\
\end{aligned}\end{equation*}
Proof is completed.

\subsection{Proof of Theorem \ref{thm:2}}\label{app:2}
Before giving the concrete analysis for Theorem \ref{thm:2}, we give a useful lemma firstly:
\begin{lem}\label{lem:2}
For any $a>0$, $b\geq0$ with $a+b\leq c<\infty$, there exists a constant $\gamma\geq0$ such that
\begin{equation*}b\leq\gamma a.\end{equation*}
\end{lem}

Since $a>0$, we have $b/a\leq c/a-1$. Because $c$ is bounded, we have $c/a-1<\infty$. Let $\gamma\geq\max\{c/a-1,0\}$, then the above inequality holds.

Now, we start to prove the result of Theorem \ref{thm:2}.

Let $\mathcal{C}_1:={\mathcal{G}}(B(s)^3)$, we define:
 \begin{equation*}\left\{\begin{aligned}
\mathring{X}\in&\mathcal{C}_2=\arg\min_{X\in\mathcal{C}_1}\|Y-\mathcal{M}X\|_F\\
\widetilde{X}\in&\mathcal{C}_3=\arg\min_{X\in\mathcal{C}_1}\|X^*-X\|_F\\
\end{aligned}\right.\end{equation*}
Let $\mathcal{S}$ represent the subset of indices sampled by $\mathcal{M}$ and $\overline{\mathcal{S}}$ represent its complement set, we have
 \begin{equation*}\begin{aligned}
\|\widetilde{X}-\mathring{X}\|_F=&\|\widetilde{X}_{\mathcal{S}}-\mathring{X}_{\mathcal{S}}+\widetilde{X}_{\overline{\mathcal{S}}}-\mathring{X}_{\overline{\mathcal{S}}}\|_F\\
\leq&\|\widetilde{X}_{\mathcal{S}}-Y-(\mathring{X}_{\mathcal{S}}-Y)\|_F+\|\widetilde{X}_{\overline{\mathcal{S}}}-\mathring{X}_{\overline{\mathcal{S}}}\|_F\\
=&\|\mathcal{M}\widetilde{X}-Y\|_F+\|\mathcal{M}\mathring{X}-Y\|_F+\|\widetilde{X}_{\overline{\mathcal{S}}}-\mathring{X}_{\overline{\mathcal{S}}}\|_F\\
\leq&2\|\mathcal{M}(\widetilde{X}-X^*)\|_F+4\|\n\|_F+\|\widetilde{X}_{\overline{\mathcal{S}}}-\mathring{X}_{\overline{\mathcal{S}}}\|_F\\
\end{aligned}\end{equation*}
where the second equality is due to $\widetilde{X}_{\mathcal{S}}=\mathcal{M}\widetilde{X}$ and $\mathring{X}_{\mathcal{S}}=\mathcal{M}\mathring{X}$, and the last inequality is due to $\|\mathcal{M}\mathring{X}-Y\|_F\leq \|\mathcal{M}\widetilde{X}-Y\|_F$.
Switch the order on both sides of the above inequality, we have
$$\|\widetilde{X}_{\mathcal{S}}-\mathring{X}_{\mathcal{S}}\|_F\leq  2\|\mathcal{M}(\widetilde{X}-X^*)\|_F+4\|\n\|_F.$$
If $\|\widetilde{X}_{{\mathcal{S}}}-\mathring{X}_{{\mathcal{S}}}\|_F=0$, then $\|Y-\mathcal{M}\widetilde{X}\|_F=\|Y-\mathcal{M}\mathring{X}\|_F$ which means that $\widetilde{X}\in \mathcal{C}_2$. There exists at least a pair $(\mathring{X},\widetilde{X})\in \mathcal{C}_2\times\mathcal{C}_3$ such that $\widetilde{X}=\mathring{X}$. Then, it holds
 \begin{equation}\label{eq:2:1}
\|\mathring{X}-X^*\|_F=\|\widetilde{X}-X^*\|_F.
\end{equation}
If $\|\widetilde{X}_{{\mathcal{S}}}-\mathring{X}_{{\mathcal{S}}}\|_F>0$, since $\mathring{\mathcal{G}}(B(s))$ is bounded, by Lemma \ref{lem:2}, there is a scalar $\gamma>0$ such that
$$\|\widetilde{X}_{{\overline{\mathcal{S}}}}-\mathring{X}_{{\overline{\mathcal{S}}}}\|_F\leq \gamma\|\widetilde{X}_{{\mathcal{S}}}-\mathring{X}_{{\mathcal{S}}}\|_F.$$
Choosing $\eta<1$ and $\eta\gamma<1$, we have
 \begin{equation*}\begin{aligned}
&\|\widetilde{X}_{\mathcal{S}}-\mathring{X}_{\mathcal{S}}\|_F\\
=&\|\widetilde{X}_{\mathcal{S}}-\mathring{X}_{\mathcal{S}}\|_F+\eta\|\widetilde{X}_{\overline{\mathcal{S}}}-\mathring{X}_{\overline{\mathcal{S}}}\|_F-\eta\|\widetilde{X}_{\overline{\mathcal{S}}}-\mathring{X}_{\overline{\mathcal{S}}}\|_F\\
\geq &\|\widetilde{X}_{\mathcal{S}}-\mathring{X}_{\mathcal{S}}\|_F+\eta\|\widetilde{X}_{\overline{\mathcal{S}}}-\mathring{X}_{\overline{\mathcal{S}}}\|_F-\eta\gamma\|\widetilde{X}_{\mathcal{S}}-\mathring{X}_{\mathcal{S}}\|_F            \\
\geq&\min\{1-\eta\gamma,\eta\}\|\widetilde{X}-\mathring{X}\|_F.
\end{aligned}\end{equation*}
With a simple derivation, it holds that
 \begin{equation}\label{eq:2:2}\begin{aligned}
\|\mathring{X}-X^*\|_F\leq&\|\widetilde{X}-X^*\|_F+\|\mathring{X}-\widetilde{X}\|_F\\
\leq&\|\widetilde{X}-X^*\|_F+\frac{2\|\mathcal{M}(\widetilde{X}-X^*)\|_F+4\|\n\|_F}{c}\\
\end{aligned}\end{equation}
where $c := \min\{1-\eta\gamma,\eta\}$. Combining (\ref{eq:2:1}) and (\ref{eq:2:2}) together, proof is completed.

\section*{Acknowledgments}
This work was supported in part by the National Key R$\&$D Program of China (2020YFA0712202, 2017YFC0108802 and 2017YFC0112903); China Postdoctoral Science Foundation under Grant 2020M682990; National Natural Science Foundation of China (61771463, 81830056, U1805261, 81971611, 61871373, 81729003, 81901736); Natural Science Foundation of Guangdong Province (2018A0303130132); Shenzhen Key Laboratory of Ultrasound Imaging and Therapy (ZDSYS20180206180631473); Shenzhen Peacock Plan Team Program (KQTD20180413181834876); Innovation and Technology Commission of the government of Hong Kong SAR (MRP/001/18X); Strategic Priority Research Program of Chinese Academy of Sciences (XDB25000000).

\bibliographystyle{model2-names.bst}\biboptions{authoryear}
\bibliography{refs}

\end{document}